\documentclass[sigconf]{acmart}
\usepackage{booktabs} 
\usepackage{color}

\setcopyright{rightsretained}

\acmDOI{10.1145/3233547.3233573}

\acmISBN{978-1-4503-5794-4/18/08}

\acmConference[ACM-BCB'18]{9th ACM International Conference on Bioinformatics, Computational Biology and Health Informatics}{August 29-September 1, 2018}{Washington, DC, USA}
\acmYear{2018} 
\copyrightyear{2018} 
\acmPrice{15.00}
\setcopyright{acmcopyright}
\acmBooktitle{ACM-BCB'18: 9th ACM International Conference on Bioinformatics, Computational Biology and Health Informatics, August 29-September 1, 2018, Washington, DC, USA}

\begin{document}
\title{Weakly Supervised Deep Learning for Thoracic  Disease Classification and Localization on Chest X-rays}

\orcid{1234-5678-9012}
\author{Chaochao Yan}
\affiliation{
  \institution{The University of Texas at Arlington}
}
\email{chaochao.yan@mavs.uta.edu}

\author{Jiawen Yao}
\affiliation{%
	\institution{The University of Texas at Arlington}
}
\email{jiawen.yao@mavs.uta.edu}

\author{Ruoyu Li}
\affiliation{%
	\institution{The University of Texas at Arlington}
}
\email{ruoyu.li@mavs.uta.edu}

\author{Zheng Xu}
\affiliation{%
	\institution{The University of Texas at Arlington}
}
\email{zheng.xu@mavs.uta.edu}

\author{Junzhou Huang}
\authornote{This work was partially supported by US National Science
Foundation IIS-1423056, CMMI-1434401, CNS-1405985,
IIS-1718853 and the NSF CAREER grant IIS-1553687.}
\affiliation{%
	\institution{The University of Texas at Arlington}
}
\affiliation{%
   \institution{Tencent AI Lab}
}
\email{jzhuang@uta.edu}

\begin{abstract}
Chest X-rays is one of the most commonly available and affordable radiological examinations in clinical practice. While detecting thoracic diseases on chest X-rays is still a challenging task for machine intelligence, due to 1) the highly varied appearance of lesion areas on X-rays from patients of different thoracic disease and 2) the shortage of accurate pixel-level annotations by radiologists for model training. Existing machine learning methods are unable to deal with the challenge that thoracic diseases usually happen in localized disease-specific areas. In this article, we propose a weakly supervised deep learning framework equipped with squeeze-and-excitation blocks, multi-map transfer, and max-min pooling for classifying thoracic diseases as well as localizing suspicious lesion regions.  The comprehensive experiments and discussions are performed on the ChestX-ray14 dataset. Both numerical and visual results have demonstrated the effectiveness of proposed model and its better performance against the state-of-the-art pipelines.
\end{abstract}

%
%



\maketitle

\fancyhead{}

\section{Introduction}

Chest X-ray imaging is currently one of the most widely available radiological examinations for screening and clinical diagnosis. However, automatic understanding of chest X-ray images is currently a technically challenging task due to the complex pathologies of different sorts of lung lesions on images. In clinical practice, the analysis and diagnosis based on chest X-rays are heavily dependent on the expertise of radiologists with at least years of professional experience. Therefore, there is a critical need of a computer-aided system that is able to automatically detect different types of thoracic diseases merely from reading patients' chest X-ray images. This is all founded on a well-designed transfer of human knowledge to machine intelligence. 

\begin{figure}[t]
  \includegraphics[width=8cm]{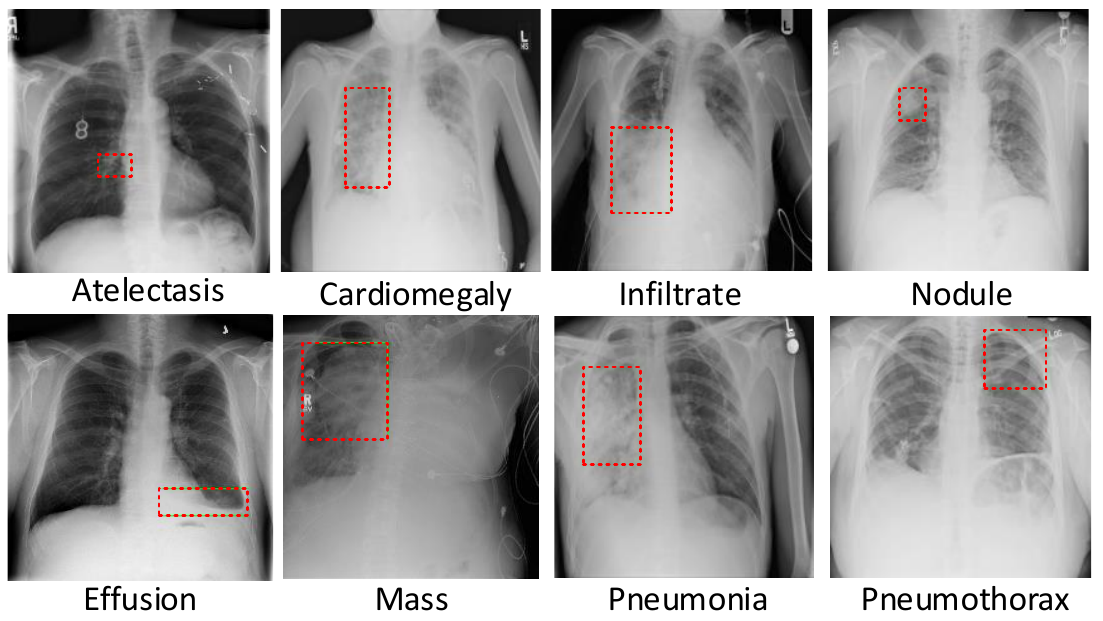}
  \caption{Examples of chest X-rays of eight thoracic diseases and associated lesion regions from ChestX-ray14~\cite{wang2017chestx}. The regions were annotated as red bounding boxes by radiologists. The bounding boxes were only used for evaluation.}~\label{fig:diseases}
\end{figure}

Since the last decade of years, as a promising technology, Medical Artificial Intelligence (Medical AI) has globally attracted interest. Especially after the emergence and fast progress of deep learning,  a revolution of computer-aided diagnosis (CAD) technique has officially started and impacted in many bio-medical applications, e.g. diabetic eye disease diagnosis~\cite{gulshan2016development},  cancer metastases detection and localization~\cite{chen15multi,li2015fastregion,liu2017detecting}, lung nodule detection~\cite{setio2017validation}, and survival analysis~\cite{zhu2017wsisa}, etc. However, introducing deep learning as solution to reading and understanding chest X-ray images is challenging due to the following reasons: 1) the visual patterns extracted from samples of different types of thoracic diseases are usually highly diverse in their appearance, sizes and locations (examples of common thoracic diseases in ChestX-ray14 dataset~\cite{wang2017chestx} are available in Fig.\ref{fig:diseases}) ; 2) retrieving massive high-quality annotations of disease, such as focal zone, on chest X-ray images is not affordable. The expenses result from both the cost of hiring experienced radiologists and the hardware requirements of collection, storage, processing of those data. Therefore, ChestX-ray14, although as the largest and most quality public chest X-rays dataset, does not provide with any pixel-wise annotations or coarse bounding boxes (example of which is in Fig.\ref{fig:diseases}) for most of chest X-ray images. Consequently, it is obvious that any machine learning  models proposed to be compatible with ChestX-ray14 dataset are required to work merely with image-level class label plus a very small amount of bounding box annotations.

Many research efforts have been made for automatic detection of thoracic diseases based on diverse data generated by chest X-ray scanning. Chapman et al.~\cite{chapman2001comparison} discussed the performance of Bayesian network and decision tree at identifying chest X-ray reports. Ye et al. ~\cite{ye2009shape} reduced false positive in classification of lung nodules on chest X-rays via introducing a weighted support vector machine (SVM) classifier. Beyond hand-crafted features, Wang et al.~\cite{wang2017chestx} concatenated the classifier to a fully convolutional network (FCN) as feature extractor in classification of thoracic diseases on images from ChestX-ray14 dataset, in which they compared four classic convolutional neural network (CNN) architectures, i.e., AlexNet~\cite{krizhevsky2012imagenet}, VGGNet~\cite{simonyan2014very}, GoogLeNet~\cite{szegedy2015going}, ResNet~\cite{he2016deep}. Later, Yao et al.~\cite{yao2017learning} investigated the hidden correlations among the 14 pathological class labels in ChestX-ray14 dataset. The most recent framework proposed by Rajpurkar et al. is CheXNet~\cite{rajpurkar2017chexnet} that fine-tuned a revised 121-layer DenseNet~\cite{huang2017densely} on ChestX-ray14 images and achieved the state-of-the-art performance on thoracic disease detection. However, those previous works typically employ single or multiple fully connected layers to densely connect and select significant features on the feature maps generated by convolutional networks. As a consequence, this architecture and its similar variants do not treat different diseases separately and thus ignore a crucial fact that those lesion areas on chest X-rays actually are disease specific. Another important issue is that many images from ChestX-ray14 contain lesion areas of more than one thoracic disease (i.e. most of images have multiple class labels). This setting is to simulate a common case that a radiologist often deals with in clinical practice. Intuitively, in their models, the classifier could be possibly confused when detecting a certain type of disease by those features extracted from the lesions belong to other diseases. Therefore, a significant improvement is expected through learning disease-discriminative features on chest X-rays. 

In this paper, we will present a novel weakly-supervised learning model to particularly overcome the aforementioned issues existing in previous works. The proposed model is able to classify thoracic diseases merely reading provided chest X-rays as well as to localize the disease regions on X-rays at pixel-level granularity. First, we harnessed the latest Fully Convolutional Network (FCN) alike model, i.e. DenseNet~\cite{huang2017densely}, as backbone network, because DenseNet has obviously shown its outstanding performance on generic image classification~\cite{huang2017densely} and semantic segmentation~\cite{jegou2017one}. Much beyond the original DenseNet, for the first time, we proposed to use the so-called "Squeeze-and-Excitation" (SE) block~\cite{hu2017squeeze}, which aims to reinforce the sensitivity of our model to subtle differences between normal and lesion regions by explicitly modeling the channel interdependencies. Moreover, we incorporated the use of multi-map transfer layers to make our network perform better to learn disease-specific features that are highly related to disease modalities, e.g. "Atelectasis" and "Nodule" on chest X-ray. The last but not the least, we realized that the max-min pooling operators~\cite{durand2017wildcat} perform better at spatially squeezing feature maps for each class of disease. 
Our major contributions in the paper are summarized as follows:
\begin{itemize}
	\item A "Squeeze-and-Excitation" block was embedded after convolution layer in DenseNet block for feature recalibration.
	\item  Concatenate stacked multi-map transfer layers to DenseNet replacing fully connected layers to mitigate the multi-label issue, which becomes crucial when labels are noisy.
	\item We incorporated the max-min pooling operator to aggregate spatial activations from multi-maps into a final prediction.
	\item Extensive experiments have been conducted to demonstrate the effectiveness of proposed methods. Our method achieved superior performance compared to the state-of-the-arts.
	\item The effectiveness of each proposed component are individually verified by experiments on ChestX-ray14 dataset.
\end{itemize}

The rest of the paper is organized as follows. Problem description and recent work on automatic detection and diagnosis techniques on chest X-rays were given in Section 1$\&$2. Then, we presented our framework details in Section 3. Experimental setup, results and discussions were in Section 4. In last section, we summarized our contributions as well as the future research highlights.




\section{Related Work}

For a long time, designing a computer-aided diagnosis platform to understand radiographs has widely attracted research interest. A well-prepared database is one of the most significant factors in successfully developing a generalizable machine learning model, especially a data-hungry deep neural network model. JSRT released a chest X-ray image set~\cite{shiraishi2000development} which contains 247 chest X-ray images including 93 normal images and 154 of those exhibited malignant and benign lung nodules. Due to the limited size of JSRT data, it is difficult to train a complex model against over-fitting. \cite{gordienko2017deep} trained a convolutional network based classifier for lung nodules classification on JSRT dataset and its improved version BSE-JSRT dataset~\cite{van2006segmentation} in which bone shadows were excluded. The Indiana chest X-ray dataset~\cite{demner2015preparing} has a mixed collection of 8,121 frontal and lateral view X-ray images together with 3,996 radiology reports contain labels from trained experts. \cite{islam2017abnormality} compared the performance of multiple state-of-the-art deep learning models on Indiana dataset for disease classification and localization of remarkable regions that contribute most to an accurate classification.

In~\cite{wang2017chestx}, a hospital-scale database, ChestX-ray14, that comprises 108,948 frontal-view of X-ray images of 32,717 individual patients was presented together with the 14 classes of image labels, each of which corresponds to a thoracic disease. Therefore, each image may have multiple labels. ChestX-ray14 is probably the largest, most quality, X-ray image dataset available publicly. It is notable that the aforementioned image labels are not directly from manual annotation by pathologists, for instead, were mined by natural language processing technique~\cite{aronson2010overview,leaman2015challenges} on associated radiaological reports. Consequently, the class labels in training set is noisy, which brings extra challenge to disease classification task.  Besides, \cite{wang2017chestx} experimentally demonstrated that those common thoracic diseases could be correctly detected or even spatial-localized via a unified weakly-supervised multi-label learning framework trained by generated noisy weak class labels. The ResNet outperformed other popular convolutional neural networks, e.g. AlexNet~\cite{krizhevsky2012imagenet}, GoogLeNet~\cite{szegedy2015going} and VGGNet-16~\cite{simonyan2014very} by rendering class-wise ROC-AUC scores e.g. 0.8141 for "Cardiomegaly". While, for some diseases like "Mass" and "Pneumonia", the scores were dramatically dragged to 0.5609 and 0.6333 respectively. This result disclosed the long-standing ignorance of the incapability of traditional CNNs on learning meaningful representations with weak supervision of noisy labels. However, the major difficulty of applying deep learning models on medical problems is the shortage of high-quality annotations by pathologist.

\begin{figure*}[t]
  \includegraphics[width=0.9\textwidth]{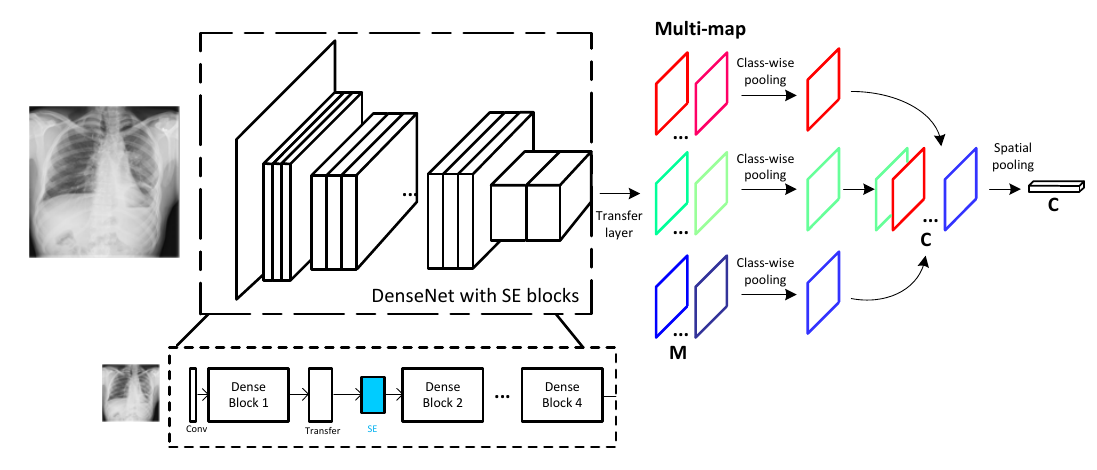}
  \caption{The Proposed Network Architecture.}~\label{fig:overview}
\end{figure*}

Shortly after ChestX-ray14 was released, \cite{rajpurkar2017chexnet} proposed a state-of-the-art CNN model named as CheXNet that consists of 121 layers. The model accepts chest X-ray images as input and outputs the probability of disease along with a heatmap which localizes the most indicative regions of disease on the input images. On the task of detecting pneumonia, the CheXNet successfully exceeded the average performance of four experienced radiologists on a subset of 420 X-ray images of pneumonia patients. However, the network in \cite{rajpurkar2017chexnet} is a variant of DenseNet~\cite{huang2017densely} without any significant modifications particularly for learning representations under a weak supervision. The network was initialized by weights pretrained on ImageNet~\cite{deng2009imagenet}, the content of which shares few in common with the images of ChestX-ray14. The lower-level representations learned on ImageNet are not guaranteed to accurately customize the shape and the contour of regions of thoracic diseases. Even though \cite{rajpurkar2017chexnet} has lifted the classification accuracy by a margin of 0.05 on ROC-AUC score, it still left quite a space for improvement. 

As mentioned in~\cite{rajpurkar2017chexnet}, the significance of comparison between CheXNet and human pathologist labeling was compromised by the fact that only the frontal view of radiographs were presented to pathologists, and it has been confirmed that there are 15\% successful diagnoses of pneumonia by pathologists mainly contributed by the lateral views, which were not available in ChestX-ray14. Consequently, a multi-view version of chest X-ray dataset - \emph{MIMIC-CXR} was presented in \cite{rubin2018large}, and based on which a dual deep convolutional network framework was naturally proposed to utilize both frontal and lateral views, if given, for disease classification. While, the network for each view was separately trained instead of weights sharing. The outputs of each network (view) were concatenated as a unified vector before a set of final fully connected layers for generating multi-class prediction. However, because of the lack of other view of radiographs in ChestX-ray14, \cite{rubin2018large} did not include a face-to-face comparison with CheXNet on the same dataset. Therefore, the actual effectiveness of introducing another relevant view of X-ray is . Moreover, the numerical results of \cite{rubin2018large} has not strongly supported the conclusion that combining more views of radiographs brings lift on recognition performance without learning the correlation between views.

As discussed above, training a classifier on X-ray images is more difficult than generic image, e.g. ImageNet, where object of interest is usually positioned in the middle of image. The lesion area of lung could be pretty small compared to the entire X-ray images. Besides, the variant condition of capturing, e.g. posture of patient, brings extra distortion and misalignment. To address these problems, \cite{guan2018diagnose} proposed an attention guided convolutional neural network (AG-CNN) to extract regions of interest (RoI) as a rough localization of lesion areas from the last convolution outputs of global network which train on raw X-ray images with class label supervision. Then, extracted RoI patches were fed to a separate local branch of CNN for learning local representation of lesion. At last, a fusion branch concatenates features generated by both global and local branches with a fine-tune with several fully-connected layers.

The ChestX-ray14 offers a very noisy class-labels and quite a few bounding boxes as ground-truth for regions of interest localization. This makes it a classic weakly supervised learning problem~\cite{crandall2006weakly,torresani2014weakly}, which is pretty common in medical areas and becoming important when developing AI in fields where expertise is expense. \cite{yao2018weakly} modeled the problem as multiple instance learning (MIL) on X-ray as a roughly-labeled bag of patches. They parameterised the Log-Sum-Exp pooling with a trainable lower-bounded adaptation (LSE-LBA) to construct illustrative saliency map at multiple resolutions.


\section{Methodology}

In this section, we will explicitly present the technical details of proposed framework. First, we illustratively discuss the advantages of DenseNet compared with other modern FCN models. Then, we individually discuss the roles of the three components that bring extra performance lift beyond DenseNet: squeeze-and-excitation block, multi-map transfer layer and max-min pooling operator. An illustration of proposed network architecture is in Fig.~\ref{fig:overview}. 

\subsection{DenseNet for Chest X-rays}
Fully convolutional network (FCN)~\cite{long2015fully} has become one of the most successful deep learning frameworks for generic image classification and segmentation tasks. In~\cite{wang2017chestx}, ResNet, a recent FCN alike model, delivered best classification accuracy on ChestX-ray8. A typical DenseNet~\cite{huang2017densely} comprises multiple densely connected convolutional layers, which improve the flow of information and gradients through the network, making it converge better and mitigating gradient vanishing issue. Therefore, in many computer vision tasks, DenseNet has shown magnificently stronger capability of representation learning than ResNet. Then \cite{rajpurkar2017chexnet} fine-tuned a DenseNet that naturally preserves spatial information throughout the network. As well as on the purpose of a fair comparison, we particularly choose the publicly available DenseNet-121 model as backbone network ~\footnote{https://github.com/pytorch/vision/blob/master/torchvision/models/densenet.py}.
As shown in Fig.~\ref{fig:overview}, the backbone of the used DenseNet consists of four consecutive dense blocks. However, original DenseNet is incapable of handling the special issues in disease classification and localization on chest X-rays. For example, disease labels of ChestX-ray14 are highly noisy since they were generated from scanning report. Given a X-ray image corresponds to multiple disease types, it is still an open question how to make data selectively contribute to multiple classification and localization tasks.

\subsection{Squeeze-and-Excitation Block in DenseNet}


In classical CNNs, it is difficult to model the interdependency between channels using convolutional filters, which are initialized and trained independently. However, the cross-channel dependency is widely existing and has been recognized as one of the major visual patterns, e.g. joint sparsity~\cite{li2015fast}. 

In between two consecutive dense blocks of DenseNet, there is a convolution-pooling operator that transforms previous activation output to a new feature space and then squeezes it to a compact spatial domain. In proposed model, we insert a so-called \emph{squeeze-and-excitation} (SE) block into the convolution-pooling operator. Particularly, we first squeeze the $C$ feature maps after convolution into a feature vector of $C$ length by spatial average-pooling. An \emph{excitation} process is to reweight feature maps by the channel-wise attention coefficients learned from the squeezed vector. The motivation is to offer a chance of cross-channel feature recalibration considering the channel interdependencies. 


\textbf{Squeeze}  
Before recalibration, we need a global statistic of each channel. Then a global squeezing is performed first by an average-pooling across entire spatial domain.
Consider $\mathbf{U} \in \mathbf{R}^{H \times W \times C}$ as transformed feature maps after convolution, where $H \times W \times C$ is the dimensionality. A \textit{squeeze} operation is to aggregate the feature maps across spatial dimensions $H \times W$ to produce a channel descriptor forming a $C$-length descriptor vector for entire $\mathbf{U}$. Assume $\mathbf{z}$ is the vector after squeezing and the $c$-th element of $\mathbf{z}$ is calculated by
\begin{align}
    z^c = \frac{1}{H\times W}\sum_{i=1}^{H}\sum_{j=1}^{W}u^c(i,j).
\end{align}
This was not possible in classical CNN in which feature maps were convolved independently by separate filter kernels and therefore the squeezing scale was constrained within reception field and the pooling was also committed locally.

\textbf{Excitation} 
To recalibrate feature maps channel-wise, we need to learn the channel weights. We employ a self-gating mechanism, which outputs channel attentions, based on the non-linear channel interdependence after passing a \emph{sigmoid} activation function $\sigma$:
\begin{align}
    \mathbf{s} =\sigma (\mathbf{W}_2 \times ReLU(\mathbf{W}_1 \times \mathbf{z}))~\label{eq:recalibration},
\end{align}
where $\mathbf{s} \in \mathbf{R}^C$ is the channel-wise attention coefficients for feature recalibration. Due to Eq~\ref{eq:recalibration}, channel coefficient $s^c$ represented the relative importance of channel $c$.
For the purpose of reducing complexity, a bottleneck structure formed by two fully connected layers parameterised by $\mathbf{W}_1 \in \mathbf{R}^{\frac{C}{r} \times C}$ and $\mathbf{W}_2 \in \mathbf{R}^{C \times \frac{C}{r}}$  ($r$ is the reduction ratio) is used in Eq~\ref{eq:recalibration} to adaptively adjust channel importance according to learning objective. The final output after SE block of channel $c$, $\Tilde{x_c}$, is obtained by re-scaling the transformed feature maps $\mathbf{U}$ with $\mathbf{s}$ by a channel-wise multiplication:
\begin{align}
\Tilde{x_c} = s^c\cdot u^c, \quad c\in \{0,\dots ,C-1\}.
\end{align}

\begin{figure}[!htb]
  \includegraphics[width=0.45\textwidth]{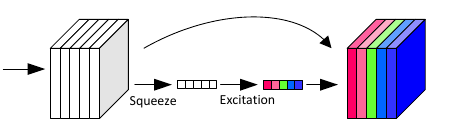}
  \caption{Illustration of a Squeeze-and-Excitation Block.}~\label{fig:se_block}
\end{figure}

The physical meaning of SE block for classification of chest X-rays comes from the hardly distinguishable illuminative contrast between lesion regions of different types of disease as well as the rest normal regions. Therefore, merely utilizing single feature map or independently processing multiple maps cannot provide enough informative features for disease classification.  The workflow of SE block is given in Fig.\ref{fig:se_block}. 

\subsection{Multi-map Layer and Max-min Pooling}
Because ChestX-ray14 offers multiple disease labels for most of X-rays, it is naturally required to perform a multi-class classification. Instead of generating a multi-hot score vector, which makes training difficult to converge, we were encouraged by good performance from introducing multi-map transfer layer, each output feature map of which corresponds to a particular disease class.

The last dense block generates feature maps with size as $w \times h \times d$. Then we concatenate to it a multi-map transfer layer. The layer encodes the activation outputs of backbone network into $M$ individual feature maps for each disease class through $1 \times1 $ convolution operation. Denote $M$ as the number of feature maps per class and $C$ as the number of classes, this transfer layer will achieve the output of size $w\times h \times MC$. When $M=1$, it is reduced to a standard classification output of $C$ classes. The modalities are learned with only image-level label and the transfer layer maintains spatial resolution. The $M$ modalities aim at specializing to different class-related visual features. 

To sufficiently utilize the provided multi-class label, we proposed a two-stage pooling layer to aggregate information on feature maps for each disease class. A standard class-wise average-pooling was first conducted to transform maps from $w \times h \times MC$ to $w \times h \times C$. As to spatial aggregation, we applied a recently proposed spatial max-min pooling~\cite{durand2017wildcat} to globally extract spatial domain information. Because we find that global minimum information is also helpful for the medical image analysis, and the global minimum regions can act as a regularizer and reduce overfitting. The global maximum and minimum pooling are linearly combined in our model:
\begin{align}
    r^c = \max_{\mathbf{h} \in \mathcal{H}_{k^{+}}}\frac{1}{k^{+}}\sum_{i,j}h_{i,j}\bar{z}_{i,j}^{c} + \alpha (\min_{\mathbf{h} \in \mathcal{H}_{k^{-}}}\frac{1}{k^{-}}\sum_{i,j}h_{i,j}\bar{z}_{i,j}^{c}),
\end{align}
where $\bar{z}^c$ is the $c$-th pooled feature map after class-wise pooling. $\mathcal{H}_k$ is the set that $\mathbf{h} \in \mathcal{H}_k$ satisfies $h_{i,j} \in \{0,1\}$ and $\sum_{i,j} h_{i,j}=k$. The max-min spatial pooling consists in selecting for each class the positive $k^{+}$ regions with the highest activations from input $\bar{z}^c$ and vice versa. The output $r^c$ is the weighted average of scores of all the selected regions. 
To generate the final positive probability, we pass $r^c$ through a sigmoid activation function. 

\subsection{Comparison with CheXNet}
The CheXNet~\cite{rajpurkar2017chexnet} is a similar model that also uses DenseNet-121 as the backbone network. It removes the last linear layer of DenseNet and adds a $1 \times 1$ convolutional layer as the transfer layer to convert the extracted 1024-channel feature maps into C-channel feature maps. To get the final $C$-dimensional output, it then uses the global maximum pooling and the sigmoid function. Compare with the CheXNet, the proposed architecture completely remove the liner layer and is fully convolutional. Our model have significant modifications particularly for learning representations under a weak supervision. We highlight the significant modifications as below:
\begin{itemize}
    \item Make the model fully convolutional by removing the linear layer. Fully convolutional architecture is suitable for spatial learning.
	\item SE blocks perform feature recalibration by weights learned from channel interdependencies, improving the representational power of CheXNet.
	\item Different from CheXNet that still only has one single feature map, we use multi-map transfer layer to encode modalities associated with each individual disease class, making our framework more capable of discriminating the appearance of multiple thoracic diseases on the same chest X-ray.
	\item To aggregate spatial scores from multi-maps into a global prediction, We incorporate a novel max-min pooling strategy which is better than the global pooling in CheXNet.
\end{itemize}

\begin{table*}[!htb]
	\caption{The comparison of AUC scores. The best AUC score in each row is displayed in bold. Note that Li et al.\cite{li2017thoracic} used extra disease location information when training the model and did not perform on official split. }
	\label{tab:aucscores}
	\begin{tabular}{ccccccc}
		\toprule
		& ChestX-ray8 ~\cite{wang2017chestx} & Yao et al.~\cite{yao2017learning} &  Li et al.~\cite{li2017thoracic} &  DNetLoc~\cite{guendel2018learning} &  CheXNet~\cite{rajpurkar2017chexnet} &   Our Method
		\\
		\midrule
		Official Split & Yes & Yes & No & Yes & Yes&  Yes\\
		Atelectasis &  0.7160 & 0.7330 & \textbf {0.8000} & 0.7670 & 0.7795 & 0.7924 \\
		Cardiomegaly &  0.8070 & 0.8580 & 0.8700 &  \textbf{0.8830} & 0.8816 &  0.8814 \\
		Effusion & 0.7840 & 0.8060 &  \textbf{0.8700}  &  0.8280 & 0.8268 & 0.8415 \\
		Infiltration &  0.6090 &  0.6750 & 0.7000 & 0.7090 & 0.6894 & \textbf{0.7095}\\    
		Mass &  0.7060 &   0.7270 & 0.8300 &    0.8210 &    0.8307 & \textbf{0.8470} \\
		Nodule & 0.6710 &   0.7780 & 0.7500 &  0.7580 & 0.7814 & \textbf{0.8105} \\
		Pneumonia & 0.6330 &  0.6900 & 0.6700 &    0.7310 & 0.7354 &  \textbf{0.7397}\\
		Pneumothorax & 0.8060 &  0.8050 &  0.8700 &  0.8460 &  0.8513 & \textbf{0.8759}\\
		Consolidation & 0.7080 &  0.7170 &  \textbf{ 0.8000 } &  0.7450 &    0.7542 &    0.7598 \\
		Edema & 0.8350 & 0.8060 & \textbf{0.8800}  &    0.8350 &   0.8496 & 0.8478  \\
		Emphysema &  0.8150 &    0.8420 &    0.9100 &    0.8950 & 0.9249 & \textbf{0.9422} \\
		Fibrosis & 0.7690 & 0.7570 &0.7800& 0.8180 & 0.8219 & \textbf{0.8326} \\
		Pleural Thickenin &  0.708 & 0.7240 &    0.7900 &    0.7610 &    0.7925 & \textbf{0.8083}
		\\
		Hernia &  0.7670 & 0.8240& 0.7700 & 0.8960 & 0.9323 &   \textbf{0.9341}
		\\ 
		\bottomrule
		Average & 0.7381 &    0.7673 &    0.8064 &    0.8066 &    0.8180 &   \textbf{0.8302}  \\
		
		\bottomrule
	\end{tabular}
\end{table*}

\section{Experiment}

\subsection{Chest X-ray Dataset}

The problem of thoracic disease classification and detection on chest X-rays has been extensively explored. Recently, Wang et al. \cite{wang2017chestx} released the largest chest X-ray dataset so far - ChestX-ray14, which collects 112,120 frontal-view chest X-ray images of 30,805 unique patients. Each radiography is labeled with one or multiple types of 14 common thorax diseases: Atelectasis, Cardiomegaly, Effusion, Infiltration, Mass, Nodule, Pneumonia, Pneumothorax, Consolidation, Edema, Emphysema, Fibrosis, Pleural Thickening and Hernia. These disease labels were mined from the associated radiological reports ($>$ 90~\% accuracy \cite{wang2017chestx}). Besides, there are 880 X-rays provided with lesion regions annotated as bounding boxes by radiologists. In our experiments, we only used disease label as ground-truth in training and evaluating the model in disease classification. We also utilized the bounding boxes only for a visual evaluation of disease region localization on X-rays.

To have a fair comparison with previous methods~\cite{wang2017chestx,yao2017learning,rajpurkar2017chexnet}, we splitted the dataset into three parts: training, validation, and evaluation, on patient level using the publicly available data split list~\cite{wang2017chestx}. There are respectively 76,524, 10,000, and 25,596 chest X-ray images for training, validation, and evaluation purposes. Since there may be multiple X-rays for each patient, split on patient level can guarantee the X-rays of the same patient be assigned to the same part. Split on image level will introduce potential over-fitting since the X-rays of the same patient can be assigned to both training and evaluation subsets.

\subsection{Experimental Setting}
Similar to ~\cite{wang2017chestx,yao2017learning,rajpurkar2017chexnet}, we formulate the Chest X-ray disease recognition as a classical multi-label classification problem. The proposed model outputs a 14-dimensional vector indicating the positive probability for each kind of listed diseases. An all-zero vector represents normal status (None of 14 listed thoracic diseases are detected). We use the standard binary cross entropy loss as objective function. ROC- AUC score (the area under the Receiver Operating Characteristic curve) are used as evaluation metric in disease classification. 

For SE blocks, we set the reduction ratio to be $16$ as suggested in~\cite{hu2017squeeze}. We set $M$ in the multi-map layer as $12$, which was experimentally proved to be an effective trade-off between the performance and the complexity. For max-min pooling, we use $k^+ = k^- = 1$ and $\alpha = 0.7$ as given in~\cite{durand2017wildcat}.
The end-to-end model was trained by Adam optimizer~\cite{Kingma2014Adam} with standard parameters ($\beta_1$ = 0.9 and $\beta_2$ = 0.99). We initialize the model using weights from the pre-trained DenseNet model, and only train the multi-map transfer layer and newly inserted Squeeze-and-Excitation layer from scratch. Following a previous work on ChestX-ray14~\cite{rajpurkar2017chexnet}, we set the batch size 16 and initial learning rate 0.0001. The learning rate will be decayed by 10 times when the validation loss plateaus for more than 5 epochs. The model of the least validation loss will be the selected classifier.

The original image size $1024 \times 1024$ is infeasible for a very deep convolutional neural network. In this paper, we resize the images to be of size $512 \times 512$ and convert single channel X-ray images into 3-channel RGB images since the pre-trained DenseNet only accepts 3-channel images as input. As ImageNet~\cite{deng2009imagenet} the pixel values in each channel were normalized. During training, we randomly crop a $448 \times 448$ sub-image from the input $512 \times 512$ image to augment the original training subset. The cropped sub-image is randomly horizontally flipped to incrementally increase the variation and the diversity of training samples. 
During the evaluation process, we use as input ten randomly cropped $448 \times 448$  sub-images (four corner crops and one central crop plus horizontally flipped version of these) for each evaluation sample, and take the average probability as the final prediction.

\subsection{Comparison with State-of-the-art Methods}

We compared the classification performance of our proposed model with previously published methods, including Wang~\cite{wang2017chestx}, Li Yao~\cite{yao2017learning}, DNetLoc~\cite{guendel2018learning}, ChexNet~\cite{rajpurkar2017chexnet} and Zhe Li~\cite{li2017thoracic}. We showed that our method achieved current state-of-the-art classification accuracy on ChestX-ray14 dataset. In the experiments, we found that different data split setup has significant influence on the model performance. However, the results of ChexNet in~\cite{rajpurkar2017chexnet} was not achieved under the official data splitting. To make a fair comparison, we implement the ChexNet and evaluate its performance with the provided official data split. It is noted that Li et al.~\cite{li2017thoracic} used extra disease location information than others in training and did not use the official split. Therefore, it is not comparable to our method as well as other state-of-the-arts approaches. Even though, we still outperformed ~\cite{li2017thoracic} in classification of 9 out of the 14 diseases.

\begin{figure*}[!htb]
  \includegraphics[width=\textwidth]{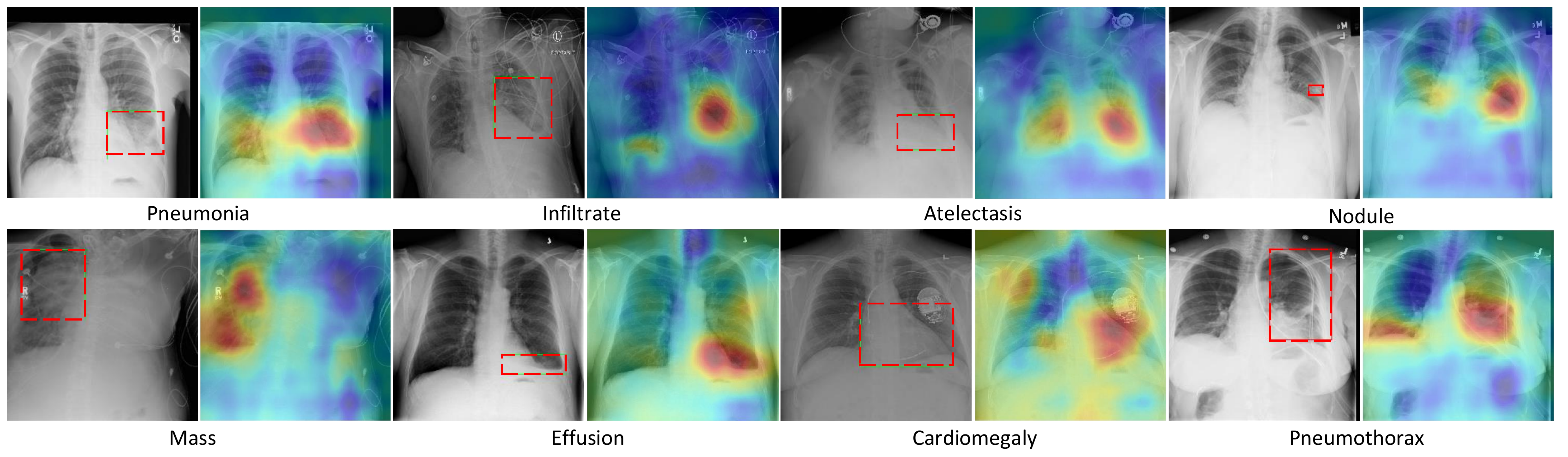}
  \caption{The proposed method localizes the areas of the X-ray that are most important for making particular pathology classification. We can see that the localized areas are very close to the corresponding bounding boxes.}~\label{fig:visual}
\end{figure*}

Numerical classification results are given in Table~\ref{tab:aucscores}. For each evaluated method, we report ROC-AUC scores for each disease class as well as the average score of all classes. Compared with previous methods, our network improves the overall performance by 2$\%$. Especially, for some challenging diseases, e.g. "Lung Nodule", the accuracy was dramatically improved by a margin of at least 3$\%$. The performance is generally improved because of the better spatial squeezing capability from the use of SE blocks and the max-min pooling operation. Moreover, for the same reason, our method can effectively handle lesion areas of different size. For example, "Cardiomegaly" and "Edema" have relatively larger pathology areas on X-rays than "Mass" and "Nodule". From Table~\ref{tab:aucscores}, it is verified that the proposed network can effectively learn decisive features from X-rays of both large and small disease areas, while others cannot.

\subsection{Localization of Lesion Regions}

In Fig.\ref{fig:visual}, we produce heat map to visualize the most indicative pathology areas on X-rays from evaluation subset, interpreting the representational power of network. Heat maps are constructed by computing the average of class-wise features after pooling along the channel dimension~\cite{guan2018diagnose}. We can see that our proposed network is able to localize lesion region on X-rays by assigning higher values than the normal. A visual evaluation has confirmed that the highlighted regions on X-rays are pretty close to ground-truth (red bounding boxes). Since our model did not use any bounding boxes in training, this has demonstrated that the proposed framework has a good interpretation ability in terms of localizing disease regions and can be widely applied in clinical practice where detailed annotations are hardly available.

\begin{table*}[!htb]
	\caption{Validation of the effectiveness of the three improvements. The best AUC score in each row is displayed in bold.}
	\label{tab:effectiveness}
	\begin{tabular}{ccccc}
		\toprule
		& Our Method & w/o SE & w/o multi-map & w/o max-min pooling
		\\
		\midrule
		Atelectasis & \textbf{0.7924} & 0.7867 & 0.7900 & 0.7784  \\
		Cardiomegaly & 0.8814 & \textbf{0.8852} & 0.8790 & 0.8762 \\
		Effusion & 0.8415 & 0.8418 & \textbf{0.8420} & 0.8392 \\
		Infiltration & \textbf{0.7095} & 0.7048 & 0.7087 & 0.6985 \\    
		Mass & \textbf{0.8470} & 0.8462 & 0.8469 & 0.8440 \\
		Nodule & 0.8105 & 0.8055 & \textbf{0.8110} & 0.8034 \\
		Pneumonia & \textbf{0.7397} & 0.7368 & 0.7364 & 0.7435 \\
		Pneumothorax & \textbf{0.8759} & 0.8738 & 0.8736 & 0.8753 \\
		Consolidation & 0.7598 & \textbf{0.7640} & 0.7586 & 0.7545 \\
		Edema & 0.8478 & 0.8464 & \textbf{0.8503} & 0.8398 \\
		Emphysema & 0.9422 & 0.9402 & \textbf{0.9436} & 0.9371 \\
		Fibrosis & \textbf{0.8326} & 0.8269 & 0.8302 & 0.8067 \\
		Pleural Thickenin & 0.7994 & \textbf{0.8059} & 0.8058 & 0.8011 \\
		Hernia & \textbf{0.9341} & 0.9330 & 0.9299 & 0.9096 \\ 
		\bottomrule
		Average & \textbf{0.8302} & 0.8279 & 0.8290 & 0.8220 \\
		
		\bottomrule
	\end{tabular}
\end{table*}

\subsection{Ablation Study}

In the section, we conduct additional ablation experiments to demonstrate the effectiveness of three proposed components in our network that respectively bring performance gains: multi-map transfer layer, max-min pooling and SE block. From Table.\ref{tab:aucscores}, CheXNet has the average AUC score as 0.8180 for all 14 diseases. The average AUC score of our method is 0.8302. This 1.2$\%$ lift demonstrated the joint effects from the three components compared to the state-of-the-art. Now we will validate the difference on performance of our model when sequentially removing each contributive component. 
It is noted that, in the next three experiments, we only changed the network structure and keep other experimental setting identical to make a fair and illustrative comparison. Results are in Table.\ref{tab:effectiveness}. 

\textbf{Effectiveness of SE Block}
In the original DenseNet architecture, the transition layer between consecutive dense blocks is simply a $1 \times 1$ convolutional layer followed by a average-pooling layer for purpose of dimension reduction. In our model, we realized that \emph{squeeze} operation will extend spatial aggregation to the entire spatial domain, which was impossible for a local pooling. Besides, \emph{excitation} process will train a parameterised reweighting on feature maps supervised by channel interdependencies, which was also not possible in previous transition layer of DenseNet. When we remove SE blocks from the network, the average AUC score drops to 0.8279 showing that our method indeed achieves performance gain by using SE blocks as they recalibrate convolutional features.

\textbf{Effectiveness of Multi-map Transfer Layer}
We use multiple feature maps for each disease in our model. In experiments, the learning of multi-map was skipped by setting $M = 1$. Consequently, the average score of revised model becomes 0.8290. As shown in Fig.~\ref{fig:overview}, the appearances of different classes of disease vary a lot in shape and color, which  supported the use of multi-map transfer layer.

\textbf{Effectiveness of Max-min Pooling}
CheXNet adopted traditional global maximum pooling which only extracts the maximum component for the whole feature map assuming the maximum component is considered to be the most informative part. However, we found that the minimum components also contribute a lot to the thoracic disease classification. Results from the Table.\ref{tab:effectiveness} validated the effectiveness of max-min pooling showing that our model would lose $1.0\%$ on ROC-AUC score when using only maximum pooling.

\section{Conclusion}
In this paper, we proposed a unified weakly-supervised deep learning framework to jointly perform thoracic disease classification and localization on chest X-rays only using noisy multi-class disease label. The advantages of proposed network are not only from the learning of disease-specific features via multi-map transfer layers, also from the cross-channel feature recalibration by \emph{sqeeuze-and-excitation} blocks in between dense blocks. Heat maps, as by-product obtained under weak supervision, further visualize the representational power of our network. This also highlights the interpretability of our model. Finally, both quantitative and qualitative results has indicated that our framework outperformed the state-of-the-arts. As to future work, we will re-investigate an accurate localization of lesion areas utilizing the limited amount of bounding boxes. 

   


\bibliographystyle{ACM-Reference-Format}
\bibliography{sample-bibliography}

\end{document}